\documentclass[11pt]{article}

\usepackage[final]{acl}

\usepackage{times}
\usepackage{latexsym}

\usepackage{amsmath}
\usepackage{cleveref}
\crefname{table}{Table}{Tables}
\crefname{figure}{Figure}{Figures}
\crefname{section}{Section}{Sections}
\crefname{appendix}{Appendix}{Appendices}
\crefname{equation}{Equation}{Equations}
\usepackage[T1]{fontenc}

\usepackage[utf8]{inputenc}

\usepackage{microtype}
\usepackage{booktabs}
\usepackage{multirow}
\usepackage{makecell}
\usepackage{pifont}
\usepackage{inconsolata}

\usepackage{graphicx}
\usepackage{colortbl}
\usepackage{xcolor}
\definecolor{heldcol}{HTML}{FFE6CC}
\definecolor{compcol}{HTML}{D5E8D4}
\definecolor{heldcol}{HTML}{FFE6CC}
\definecolor{traincompcol}{HTML}{D5E8D4}
\definecolor{testcompcol}{HTML}{FFFFFF}

\usepackage{pgf}

\definecolor{grad0}{RGB}{63,28,77}     
\definecolor{grad1}{RGB}{84,42,98}     
\definecolor{grad2}{RGB}{104,58,112}   
\definecolor{grad3}{RGB}{120,80,116}   
\definecolor{grad4}{RGB}{130,108,114}  
\definecolor{grad5}{RGB}{132,140,110}  
\definecolor{grad6}{RGB}{122,164,100}  
\definecolor{grad7}{RGB}{104,184,94}   
\definecolor{grad8}{RGB}{120,204,110}  
\definecolor{grad9}{RGB}{150,222,140}  

\newcommand{\valcell}[1]{%
  \ifdim #1pt<0.1pt \cellcolor{grad0}\textcolor{white}{#1}\else
  \ifdim #1pt<0.2pt \cellcolor{grad1}\textcolor{white}{#1}\else
  \ifdim #1pt<0.3pt \cellcolor{grad2}\textcolor{white}{#1}\else
  \ifdim #1pt<0.4pt \cellcolor{grad3}\textcolor{white}{#1}\else
  \ifdim #1pt<0.5pt \cellcolor{grad4}\textcolor{white}{#1}\else
  \ifdim #1pt<0.6pt \cellcolor{grad5}{#1}\else
  \ifdim #1pt<0.7pt \cellcolor{grad6}{#1}\else
  \ifdim #1pt<0.8pt \cellcolor{grad7}{#1}\else
  \ifdim #1pt<0.9pt \cellcolor{grad8}{#1}\else
  \cellcolor{grad9}{#1}\fi\fi\fi\fi\fi\fi\fi\fi\fi}
\newcommand{\cmark}{\textcolor{teal}{\ding{51}}}
\newcommand{\xmark}{\textcolor{red}{\ding{55}}}

\title{Multi-Hop Knowledge Composition is Bound by Pretraining Exposure}

\author{
  \textbf{Yannis Karmim\textsuperscript{1,2,3}},
  \textbf{Luis Mart\'{i}\textsuperscript{2}},
  \textbf{Djamé Seddah\textsuperscript{1}},
  \textbf{Valentin Barriere\textsuperscript{3,4}}
\\
\\
  \textsuperscript{1}Inria, Paris, France,
  \textsuperscript{2}Inria Chile, Las Condes, Chile,\\
  \textsuperscript{3}Dept. of Computer Science, Univ. de Chile, \textsuperscript{4} CENIA, Chile
\\
  \small{
    \textbf{Correspondence:} \href{mailto:yannis.karmim@inria.fr}{yannis.karmim@inria.fr}
  }
}

\begin{document}
\maketitle

\begin{abstract}
Large Language Models fail at implicit multi-hop reasoning (or latent composition):
a model answers \textit{"When was $X$ born?"} and
\textit{"Who is $Y$'s closest friend?"} correctly but fails on
\textit{"When was $Y$'s closest friend born?"} in a single forward
pass, even when both facts are perfectly memorized and individually
retrievable. 
We study this failure in a controlled natural language
setting with a strict separation between individuals exposed to
compositional contexts during pretraining and those that never appear
in any such context.  We confirm that compositional failure persists
even at 97\% 1-hop accuracy, establishing the gap as a pretraining
failure rather than a knowledge absence. We propose and test nine
data-centric augmentation formats and find that compositional
pretraining transfers to unseen questions for exposed individuals, but
never to individuals absent from compositional pretraining, suggesting
that exposure to compositional contexts during pretraining is a
necessary condition for implicit multi-hop reasoning. Our code is \href{https://github.com/ykrmm/composition-bound}{available}.
\end{abstract}

\section{Introduction}

\begin{figure*}[!ht]
   \centering
   \includegraphics[width=0.95\linewidth]{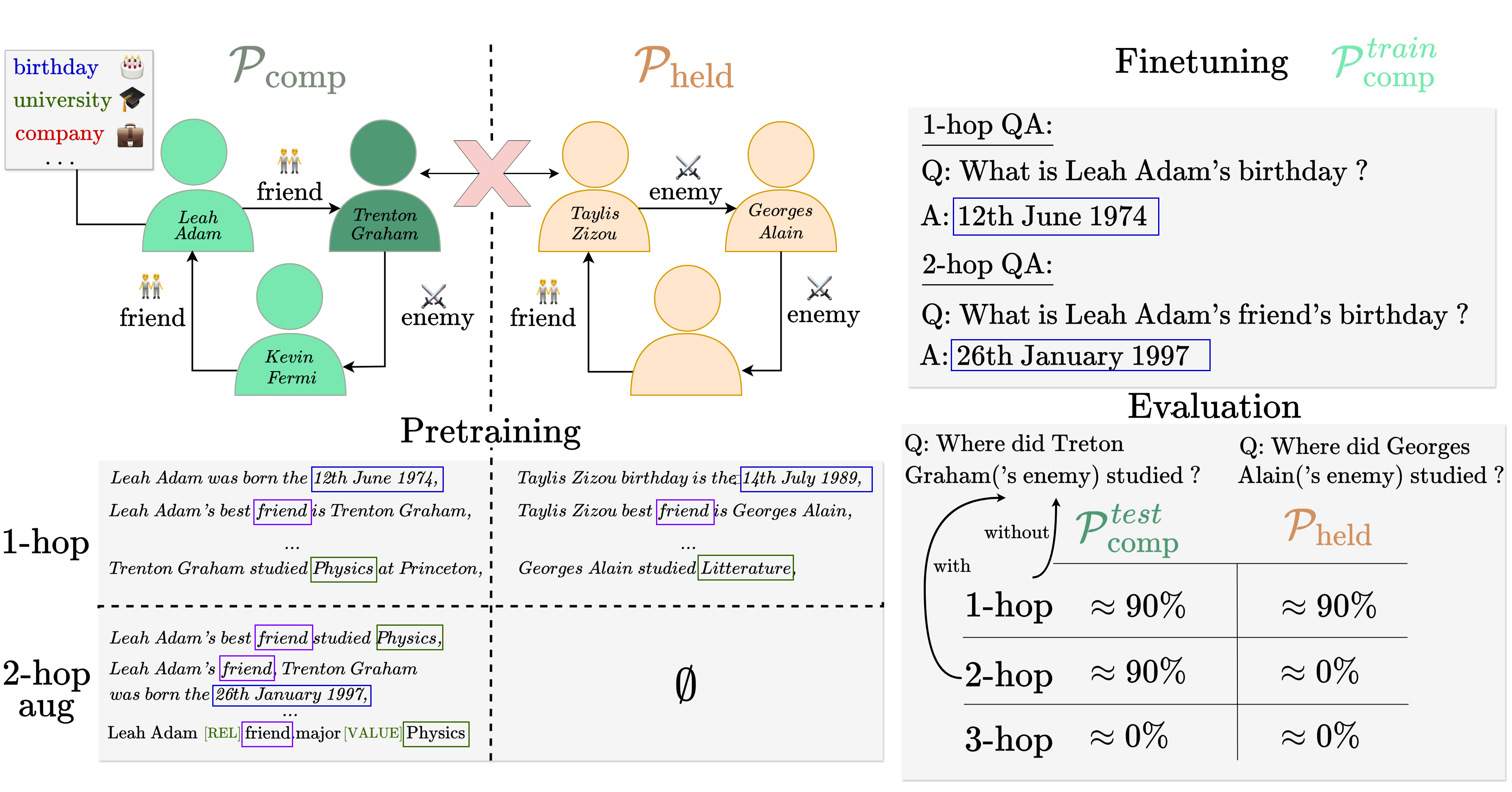}
       \caption{\textbf{Overview.} \textit{(i)} We study the 
compositionality gap in a controlled natural language setting, extending the synthetic biography framework of 
\citet{allen-zhu_physics_2023-1} with \texttt{friend} 
and \texttt{enemy} relations to enable implicit reasoning on multi-hop QA. 
\textit{(ii)} We pretrain on all $1$-hop biographies ($\mathcal{P}_\text{comp}+\mathcal{P}_\text{held}$) and perform 2-hop data augmentation only on $\mathcal{P}_\text{comp}$.
\textit{(iii)}
We then perform multi-hop QA fine-tuning on $\mathcal{P}_\text{comp}^{\text{train}}$, which transfers well to $\mathcal{P}_\text{comp}^{\text{test}}$ depending on the augmentation strategy, but never, under any tested condition, to $\mathcal{P}_{\text{held}}$.}
   \label{fig:intro}
\end{figure*}

Large language models (LLMs) store and retrieve 
factual knowledge with surprising 
fidelity~\cite{allen-zhu_physics_2023-1, 
petroni-etal-2019-language}, yet struggle on the 
simplest form of \textbf{implicit} multi-hop reasoning. A 
model that correctly answers \textit{"When was 
Thierry born?"} and \textit{"Who is Zinedine's 
closest friend?"} may fail on \textit{"When was 
Zinedine's closest friend born?"}, a query where 
the bridge entity \textit{Thierry} is absent, even 
when both constituent facts are individually 
extractable. This failure, 
the compositionality gap \cite{press_measuring_2023}, 
is robust to model scale and 
persists across architectures and recent frontier models like Claude, Gemini or Mistral \cite{xu_large_2024,yang-etal-2024-large-language-models,yang-etal-2025-large}. Existing solutions 
externalize composition at inference time via 
chain-of-thought, decomposition, or model patching 
\cite{wei_chain--thought_2023, press_measuring_2023, 
biran_hopping_2024}, leaving open whether 
compositional reasoning can emerge from pretraining 
itself. On proprietary models, the cause remains 
undiagnosable as pretraining data is undisclosed. 
\citet{allen-zhu_physics_2023-1} show that knowledge memorization and extraction are distinct in  controlled biography settings but study only  single-hop retrieval. \citet{ye2026transformers}  show that second-hop generalization requires query-level exposure but rely on symbolic tokens (\citealt{wang_grokking_2024}) without any augmentation.

In this short paper, we ask whether data-centric 
pretraining augmentations can induce implicit 
multi-hop composition in a controlled natural 
language setting. We extend the synthetic biography framework of \citet{allen-zhu_physics_2023-1} with inter-individual relations and partition individuals into a composed population and a held-out population 
restricted to atomic 1-hop biographies only. We then finetune on a subset of exposed individuals and evaluate on the remainder, separating transfer across unseen questions from transfer across individuals never seen in any compositional context during pretraining. 
Compositional augmentation substantially improves 
multi-hop QA on unseen questions for exposed 
individuals, but never transfers to those absent 
from compositional pretraining, establishing 
pretraining exposure as a necessary condition for 
implicit multi-hop composition. Our setup is 
summarized in \Cref{fig:intro}.

Our contributions are as follows: 
\textit{(i)} we release an open reproduction of the biography framework~\cite{allen-zhu_physics_2023-1} extended with inter-individual relations and multi-hop QA, 
\textit{(ii)} we confirm in natural language that 
a model at 97\% 1-hop accuracy scores near 0\% on 
2-hop queries, establishing the compositionality 
gap as a pretraining failure, \textit{(iii)} across 9 augmentation formats, compositional pretraining transfers to unseen questions for exposed individuals but never to the held-out population.

\vspace{-0.3em}
\section{Related Work}
\vspace{-0.3em}
\label{sec:rw}

\paragraph{Compositionality gap and implicit reasoning.}
Language models struggle to implicitly compose 
individually retrievable facts within a single 
forward pass \citep{press_measuring_2023}, and 
scaling can worsen this by favoring memorization 
shortcuts \citep{wang_larger_2025}. This setting 
is particularly challenging because the required 
reasoning is structurally simple, yet composition 
must occur entirely in parametric knowledge without external context. Compositional failures in parametric settings do not extend to in-context learning \citep{10.5555/3737916.3740788, 
allen-zhu_physics_2023}, motivating separate study of these regimes. Existing approaches externalize composition through chain-of-thought, decomposition, or external modules \citep{wei_chain--thought_2023, 
press_measuring_2023, chen_skills--context_2024}. \cite{yao-etal-2025-language-models} show that language models can learn In Distribution (ID) implicit k-hop reasoning but the required training data grows exponentially in $k$.
\begin{table*}[!h]
    \centering
    \small
    \setlength{\tabcolsep}{4pt}
    \resizebox{0.80\linewidth}{!}{
    \begin{tabular}{clccccc}
    \toprule
    \multirow{2}{*}{\textbf{\# Exp.}} & 
    \multirow{2}{*}{\textbf{Format}} & 
    \multicolumn{2}{c}{\textbf{Natural Language (NL)}} & 
    \multicolumn{3}{c}{\textbf{RDF}} \\
    \cmidrule(lr){3-4} \cmidrule(lr){5-7}
    & & 
    {\scriptsize \textbf{Explicit 2-hop}} & 
    {\scriptsize \textbf{Implicit 2-hop}} & 
    {\scriptsize \textbf{1-hop}} & 
    {\scriptsize \textbf{Explicit 2-hop}} & 
    {\scriptsize \textbf{Implicit 2-hop}} \\
    \midrule
    Exp. 0 & Baseline (no augmentation) 
      & \xmark & \xmark & \xmark & \xmark & \xmark \\
    Exp. 1 & PT RDF 
      & \xmark & \xmark & \cmark & \xmark & \xmark \\
    Exp. 2 & PT 2-hop implicit NL 
      & \xmark & \cmark & \xmark & \xmark & \xmark \\
    Exp. 3 & PT 2-hop explicit NL 
      & \cmark & \xmark & \xmark & \xmark & \xmark \\
    Exp. 4 & PT 2-hop implicit RDF 
      & \xmark & \xmark & \xmark & \xmark & \cmark \\
    Exp. 5 & PT 2-hop explicit RDF 
      & \xmark & \xmark & \xmark & \cmark & \xmark \\
    Exp. 6 & PT 2-hop implicit + explicit RDF 
      & \xmark & \xmark & \xmark & \cmark & \cmark \\
    Exp. 7 & PT 2-hop implicit + explicit NL 
      & \cmark & \cmark & \xmark & \xmark & \xmark \\
    Exp. 8 & PT 2-hop implicit + explicit NL-RDF
      & \cmark & \cmark & \xmark & \cmark & \cmark \\
    Exp. 9 & All formats 
      & \cmark & \cmark & \cmark & \cmark & \cmark \\
    \bottomrule
    \end{tabular}
    }
    \caption{
Data augmentation strategy for pretraining (PT) over $\mathcal{P}_{\text{comp}}$. All conditions include natural-language 1-hop biographies with \texttt{multi5p-permute} augmentation for every individuals. Explicit settings include the bridge entity.
}
    \label{tab:augmentations}
\end{table*}

\vspace{-0.7em}

\paragraph{Mechanistic approaches.}
Mechanistic studies trace compositional failure to 
insufficient propagation of intermediate entities 
\citep{li_understanding_2024, biran_hopping_2024, 
hou_towards_2023}. \citet{biran_hopping_2024} filter 
behavioral shortcuts at evaluation time, but 
co-occurrence of constituent facts during pretraining 
remains uncontrolled, leaving open whether success 
reflects genuine composition or joint memorization. 
Controlled studies isolate the phenomenon but remain 
limited: \citet{ye2026transformers} use symbolic 
tokens without augmentation, while
\citet{wang_grokking_2024} find implicit reasoning 
emerges via grokking with systematic Out Of Distribution (OOD) 
failure. \citet{balesni_two-hop_2024} show that 
facts learned in separate documents fail to compose. \citet{10.5555/3692070.3694117} show that augmentingpretraining with random-walk paths over knowledge graphs improves multi-hop reasoning, but without a strict separation between training and test entities.
We extend this line of work to natural language, 
with strict co-occurrence control and a systematic evaluation of data-centric pretraining interventions.
\vspace{-0.3em}
\section{Controlled Multi-Hop Setting}
\label{sec:setup}
\vspace{-0.5em}
We present here our controlled natural language setting to study implicit multi-hop knowledge reasoning. We describe here successively the dataset and population partition, the pretraining setup, our new data augmentation strategy, as well as the finetuning and evaluation protocol.

\textbf{Dataset construction.}
We build on the synthetic biography framework of 
\citet{allen-zhu_physics_2023-1}: $N=100$K 
individuals, each described by six attributes 
(\texttt{birthday}, \texttt{birthcity}, 
\texttt{university}, \texttt{major}, 
\texttt{company}, \texttt{workcity}). We add a 
unique directional \texttt{friend} and \texttt{enemy} 
relation per individual, enabling multi-hop queries 
such as \textit{What is the birthday of X's 
friend's enemy?}, where the bridge entity $Y=X.\text{friend}$ is absent from the query. We partition individuals into 
$\mathcal{P}_{\text{comp}}$, exposed to 
compositional pretraining contexts, and 
$\mathcal{P}_{\text{held}}$, restricted to atomic 
biographies only. Relations are defined exclusively 
within each population ensuring that $\mathcal{P}_{\text{held}}$ 
individuals never appear as intermediate entities 
in any compositional chain.

\textbf{Pretraining setup.}
We pretrain GPT-2 Small and evaluate scalability 
on GPT-2 Medium and Large~\cite{radford2019language}, 
all trained from scratch with rotary positional 
embeddings~\cite{su_roformer_2023}. Training batches mix atomic biographies from all $N$ individuals with compositional augmentation sequences drawn exclusively from $\mathcal{P}_{\text{comp}}$. Atomic biographies follow the \texttt{multi5p-permute} format of \citet{allen-zhu_physics_2023-1}: five paraphrased versions per individual with permuted attribute order, shown to be necessary for reliable 1-hop extraction. Dataset statistics and token 
counts are reported in \Cref{tab:token_count}.

\textbf{Augmentation strategy.}
We study 9 augmentation conditions varying two 
axes: data format (natural language vs.\ Resource Description Framework (RDF)) and 
bridge verbalization (explicit vs.\ implicit), 
summarized in \Cref{tab:augmentations}. Prior work 
uses only implicit symbolic contexts 
\citep{ye2026transformers}. We additionally 
hypothesize that explicitly verbalizing the bridge 
entity may align representations between explicit 
and implicit compositions, making the intermediate 
entity easier to retrieve during inference. Since 
composition becomes substantially easier once the 
bridge entity is identified \citep{biran_hopping_2024}, 
explicit and implicit formulations may reinforce a 
shared compositional representation. We use the RDF format to
isolate relational structure from lexical variation. 
Mix ratios and format examples are in 
\Cref{app:ratio} and \Cref{tab:format_examples}.

\textbf{Finetuning and evaluation task.}
After pretraining we finetune on 75\% of individuals composed during pretrain $\mathcal{P}^{\text{train}}_{\text{comp}}$
We evaluate using greedy-decoding on 1-hop, 2-hop, and 3-hop reasoning in a single forward pass without access to intermediate entities in context. We measure performance using exact-match accuracy, we report the average performance over five runs with different seeds. All training and optimization configurations are provided in \Cref{app:train}.
\section{Empirical Results}
We evaluate multi-hop QA accuracy across populations and augmentation conditions. We diagnose the compositional gap under standard training, then systematically evaluate the 9 data augmentations. Additional experiments are presented in \Cref{app:exp}.

\begin{table*}[!ht]
\centering
\small
\setlength{\tabcolsep}{4pt}
\resizebox{0.80\linewidth}{!}{
\begin{tabular}{cccccccc}
\toprule
\multirow{2}{*}{\textbf{\# Exp.}}& \multirow{2}{*}{\textbf{Data augmentation}} &
\multicolumn{3}{c}{$\mathcal{P}^{\text{test}}_{\text{comp}}$} &
\multicolumn{3}{c}{$\mathcal{P}_{\text{held}}$} \\
\cmidrule(lr){3-5} \cmidrule(lr){6-8}
& & \textbf{1-hop} & \textbf{2-hop} & \textbf{3-hop} & \textbf{1-hop} & \textbf{2-hop} & \textbf{3-hop} \\
\midrule
Exp. 1 & PT RDF & \valcell{0.97} & \valcell{0.01} & \valcell{0.01} & \valcell{0.97} & \valcell{0.01} & \valcell{0.01} \\
Exp. 2 & PT 2-hop implicit NL & \valcell{0.83} & \valcell{0.54} & \valcell{0.01} & \valcell{0.75} & \valcell{0.01} & \valcell{0.01} \\
Exp. 3 & PT 2-hop explicit NL & \valcell{0.96} & \valcell{0.02} & \valcell{0.02} & \valcell{0.89} & \valcell{0.02} & \valcell{0.02} \\
Exp. 4 & PT 2-hop implicit RDF & \valcell{0.96} & \valcell{0.71} & \valcell{0.01} & \valcell{0.40} & \valcell{0.01} & \valcell{0.01} \\
Exp. 5 & PT 2-hop explicit RDF & \valcell{0.97} & \valcell{0.02} & \valcell{0.02} & \valcell{0.37} & \valcell{0.02} & \valcell{0.02} \\
Exp. 6 & PT 2-hop implicit + explicit RDF & \valcell{0.98} & \valcell{0.69} & \valcell{0.01} & \valcell{0.48} & \valcell{0.01} & \valcell{0.01} \\
Exp. 7 & PT 2-hop implicit + explicit NL & \valcell{0.89} & \valcell{0.68} & \valcell{0.01} & \valcell{0.79} & \valcell{0.01} & \valcell{0.01} \\
Exp. 8 & PT 2-hop implicit + explicit NL-RDF & \valcell{0.99} & \valcell{0.76} & \valcell{0.01} & \valcell{0.82} & \valcell{0.01} & \valcell{0.01} \\
Exp. 9 & All formats & \valcell{0.98} & \valcell{0.71} & \valcell{0.01} & \valcell{0.80} & \valcell{0.01} & \valcell{0.01} \\
\bottomrule
\end{tabular}
}
\caption{\textbf{GPT-2 results across augmentation
conditions.} Exact-match accuracy for 1, 2 and
3-hop queries on $\mathcal{P}^{\text{test}}_{\text{comp}}$
(unseen at finetuning) and $\mathcal{P}_{\text{held}}$
(never compositionally exposed). Each value is the
mean over 5 finetuning and evaluation seeds;
standard deviations are below $0.006$ for all cells.
Cell color encodes accuracy from low to high.
2-hop accuracy on $\mathcal{P}_{\text{held}}$ is at
chance across all conditions.}
\label{tab:aug_results}
\end{table*}

\begin{table*}[!ht]
\centering
\small
\setlength{\tabcolsep}{4pt}
\resizebox{0.65\linewidth}{!}{
\begin{tabular}{cccc|ccc|ccc}
\toprule
\multirow{2}{*}{\#} &
\multicolumn{3}{c}{$\mathcal{P}^{\text{train}}_{\text{comp}}$} &
\multicolumn{3}{c}{$\mathcal{P}^{\text{test}}_{\text{comp}}$} &
\multicolumn{3}{c}{$\mathcal{P}_{\text{held}}$} \\
\cmidrule(lr){2-4} \cmidrule(lr){5-7} \cmidrule(lr){8-10}
& 1-hop & 2-hop & 3-hop & 1-hop & 2-hop & 3-hop &
1-hop & 2-hop & 3-hop \\
\midrule
Exp. 0 & \valcell{0.99} & \valcell{0.02} & \valcell{0.02} & \valcell{0.99} & \valcell{0.02} & \valcell{0.02} & \valcell{0.95} & \valcell{0.02} & \valcell{0.02} \\
\midrule
Exp. 1 & \valcell{0.99} & \valcell{0.00} & \valcell{0.00} & \valcell{0.99} & \valcell{0.00} & \valcell{0.00} & \valcell{0.97} & \valcell{0.00} & \valcell{0.00} \\
Exp. 2 & \valcell{0.99} & \valcell{0.76} & \valcell{0.02} & \valcell{0.95} & \valcell{0.69} & \valcell{0.02} & \valcell{0.91} & \valcell{0.02} & \valcell{0.02} \\
Exp. 3 & \valcell{1.00} & \valcell{0.02} & \valcell{0.02} & \valcell{0.97} & \valcell{0.02} & \valcell{0.02} & \valcell{0.92} & \valcell{0.02} & \valcell{0.02} \\
Exp. 4 & \valcell{0.99} & \valcell{0.86} & \valcell{0.01} & \valcell{0.98} & \valcell{0.77} & \valcell{0.01} & \valcell{0.99} & \valcell{0.01} & \valcell{0.01} \\
Exp. 5 & \valcell{1.00} & \valcell{0.02} & \valcell{0.02} & \valcell{0.99} & \valcell{0.02} & \valcell{0.02} & \valcell{0.44} & \valcell{0.02} & \valcell{0.02} \\
Exp. 6 & \valcell{0.99} & \valcell{0.84} & \valcell{0.01} & \valcell{0.99} & \valcell{0.74} & \valcell{0.01} & \valcell{0.56} & \valcell{0.01} & \valcell{0.01} \\
Exp. 7 & \valcell{0.99} & \valcell{0.69} & \valcell{0.01} & \valcell{0.97} & \valcell{0.58} & \valcell{0.01} & \valcell{0.95} & \valcell{0.01} & \valcell{0.01} \\
Exp. 8 & \valcell{1.00} & \valcell{0.89} & \valcell{0.01} & \valcell{0.99} & \valcell{0.78} & \valcell{0.01} & \valcell{0.87} & \valcell{0.01} & \valcell{0.01} \\
Exp. 9 & \valcell{1.00} & \valcell{0.87} & \valcell{0.01} & \valcell{0.99} & \valcell{0.77} & \valcell{0.01} & \valcell{0.94} & \valcell{0.01} & \valcell{0.01} \\
\bottomrule
\end{tabular}
}
\caption{\textbf{Llama-3 architecture: full results across augmentation conditions.} Exact-match accuracy for 1-, 2- and 3-hop queries on $\mathcal{P}^{\text{train}}_{\text{comp}}$,
$\mathcal{P}^{\text{test}}_{\text{comp}}$ and $\mathcal{P}_{\text{held}}$, under greedy decoding. Each cell is the mean over 5 finetuning seeds; the standard deviation is below $0.006$ everywhere, so we report the mean only.}

\label{tab:aug_results_full_llama}
\end{table*}

\textbf{(i) Compositional gap diagnosis.}
Under 1-hop finetuning (Table~\ref{tab:baseline}), 
models reach near-perfect 1-hop accuracy on both 
populations, confirming \texttt{multi5p} resolves 
1-hop extraction \cite{allen-zhu_physics_2023-1}, 
but 2-hop and 3-hop remain at chance. Adding 2-hop 
questions to finetuning marginally improves 
$\mathcal{P}_{\text{comp}}$ (0.08) but not 
$\mathcal{P}_{\text{held}}$, despite perfectly 
memorized 1-hop facts. Our LoRA sweep 
($r_{\text{qv}}\!\in\!\{8,16,32\}$, 
$r_{\text{emb}}\!\in\!\{32,64,128\}$) yields no 
improvement, and full finetuning converges on 
$\mathcal{P}_{\text{comp}}$ at the cost of 
catastrophic forgetting on $\mathcal{P}_{\text{held}}$ 
(\Cref{tab:forget}), indicating that compositional information is absent from pretrained representations. 
This pattern persists across model scales (Table~\ref{tab:scale}), as well as the guarantee that we are not in an under-training scheme (\Cref{tab:token_count}), this is  consistent with findings that the compositionality gap is robust to model scale \citep{press_measuring_2023}. The 
compositional gap is a pretraining failure, not a 
capacity limit.

\begin{table}[!h]
\centering
\small
\begin{tabular}{llccc}
\toprule
\textbf{Finetuning} & \textbf{Pop.} & \textbf{1-hop} & \textbf{2-hop} & \textbf{3-hop} \\
\midrule
1-hop only & $\mathcal{P}_{\text{comp}}$ & \valcell{1.00} & \valcell{0.01} & \valcell{0.01} \\
           & $\mathcal{P}_{\text{held}}$ & \valcell{0.97} & \valcell{0.01} & \valcell{0.01} \\
\midrule
1+2-hop    & $\mathcal{P}_{\text{comp}}$ & \valcell{1.00} & \valcell{0.08} & \valcell{0.01} \\
           & $\mathcal{P}_{\text{held}}$ & \valcell{0.93} & \valcell{0.01} & \valcell{0.01} \\
\bottomrule
\end{tabular}
\caption{\textbf{Baseline LoRA finetuning.} Near-perfect 1-hop accuracy on both populations under both regimes. 2-hop stays marginal on $\mathcal{P}_{\text{comp}}$ and at chance on $\mathcal{P}_{\text{held}}$. }
\label{tab:baseline}
\end{table}

\begin{table}[!h]
\centering
\small
\begin{tabular}{lccc}
\toprule
\textbf{Model} & \textbf{1-hop} & \textbf{2-hop} & \textbf{3-hop} \\
\midrule
GPT-2 small (124M)  & \valcell{0.93} & \valcell{0.01} & \valcell{0.01} \\
GPT-2 medium (354M) & \valcell{0.94} & \valcell{0.01} & \valcell{0.01} \\
GPT-2 large (774M)  & \valcell{0.94} & \valcell{0.01} & \valcell{0.01} \\
\bottomrule
\end{tabular}

\caption{\textbf{Ablation: scale analysis} on $\mathcal{P}_{\text{held}}$ under 1+2-hop finetuning. The compositional gap persists regardless of model size.}
\vspace{-1.5em}
\label{tab:scale}
\end{table}

\pagebreak[4]

\textbf{(ii) Data-centric augmentation.}
\Cref{tab:aug_results} reports accuracy of a GPT-2 architecture across the 9 augmentation conditions, highlighting four key dynamics. First, pretraining exposure is a strict prerequisite for multi-hop composition. 2-hop accuracy on $\mathcal{P}_{\text{held}}$ remains at chance in exact match (0.01--0.02) across all conditions, showing that no data-centric augmentation compensates for missing exposure, consistent with symbolic findings on OOD triplets~\cite{ye2026transformers}. Second, implicit augmentation consistently outperforms explicit augmentation in isolation. Explicit-only setups (Exp.~3, 5) remain at chance level (0.02), matching the baseline, on $\mathcal{P}^{\text{test}}_{\text{comp}}$, whereas implicit NL (Exp.~2) reaches 0.54 and implicit RDF (Exp.~4) 0.71. This suggests that explicit supervision encourages direct association learning, while implicit supervision better matches the inference setting where the bridge entity is absent. Explicit augmentation becomes beneficial when combined with implicit formats: implicit+explicit RDF (Exp.~6) reaches 0.69 and NL+RDF (Exp.~8) peaks at 0.76 on $\mathcal{P}^{\text{test}}_{\text{comp}}$, suggesting that mixing explicit and implicit signals improves representation alignment across NL and RDF formats. Notably, implicit RDF alone (Exp.~4, 0.71) approaches this performance at a fraction of the data cost (\Cref{tab:token_count}), suggesting structured triples as a lightweight and effective format for compositional pretraining augmentation.
Transfer to $\mathcal{P}_{\text{held}}$ 
remains zero under all conditions. To verify that the compositionality gap is not an artifact of the GPT-2 architecture, \cref{tab:aug_results_full_llama} demonstrates that our findings hold when replicating the pipeline on a parameter-matched Llama-style model. Comprehensive details regarding the architectural modifications and training setup are provided in \cref{app:llama}. Finally, gains on $\mathcal{P}^{\text{test}}_{\text{comp}}$ come at a cost to 1-hop retention (Exp.~4: 0.40), likely due to dilution of atomic biographies in training batches. We fix the atomic/compositional ratio at 30/70 following prior compositional supervision regimes~\citep{ye2026transformers,wang_grokking_2024,yao-etal-2025-language-models}. Mixing details in \Cref{app:ratio}.

\section{Conclusion}

We demonstrate that data-centric pretraining 
augmentation can induce multi-hop composition for 
exposed individuals, but never transfers to 
individuals absent from compositional pretraining 
contexts, confirming the fundamental limit of this 
approach beyond symbolic settings~\cite{ye2026transformers}. 
Models fail on compositions of length two for unexposed 
individuals, suggesting augmentation induces 
entity-specific associations rather than a 
reusable composition logic. We also show that mixing 
graph with natural language enables composition 
over knowledge graph entities, consistent with recent 
efforts to unify LLMs and knowledge 
graphs~\citep{pan2024unifying}. We release code, data, 
and our reproduction of \citet{allen-zhu_physics_2023-1} 
to support future investigation into training objectives 
or architectural modifications that may overcome this limit.

\section*{Limitations}
Due to limitied computational ressources our augmentation experiments use GPT-2 small to 
large. We argue scale would not change our central claim, as the compositionality gap persists up to 175B parameters \citep{press_measuring_2023} and 
our own baseline confirms near-zero 2-hop accuracy from 124M to 774M.  More fundamentally, our claim concerns exposure, not capacity: $\mathcal{P}_{\text{held}}$ 
fails because its individuals never appear in a 
compositional context during pretraining, and no 
amount of parameters supplies a missing training 
signal. However, LLMs trained on long-context 
naturally co-occurring text raise a separate 
question: whether naturally co-occurring related facts in long-context pretraining corpora could implicitly substitute for explicit compositional supervision remains an open question our synthetic setting cannot address.

We focus on data-centric interventions under 
standard pretraining and supervised finetuning, 
and do not evaluate alternative training objectives such as reinforcement learning~\cite{hatamizadeh2026rlp} or knowledge distillation~\cite{yu2024distilling}, which may offer complementary solutions to the compositional gap. We evaluate a single mixing ratio per condition, argued to be a conservative choice in \Cref{app:ratio}. Our setting is fully synthetic, and generalization to naturally occurring corpora remains untested. We restrict evaluation to two relation types (\texttt{friend}, \texttt{enemy}), leaving open whether findings extend to richer relational structures.

\section*{Acknowledgments}
The first author was fully funded by the INRIA's {\em Direction des Relations Internationales} and conducted this work during his stay at Universidad de Chile and Inria Chile. This work has received partial funding from Djamé Seddah's chairs in the PRAIRIE-PSAI, funded by the French national agency ANR, as part of the "France 2030" strategy under the reference ANR-23-IACL-0008. 
This work was partially financed with the grant U-INICIA 2024 from the Vicerrectoría de Investigación y Desarrollo (VID) number UI-011/24 "Estudios de sesgos sociales en modelos de lenguajes largos", and by the ANID fondecyt grant 11251024 "Multimodal Argumentation Mining in Groups Assissted by an Embodied Conversational Agent", and by the Franco-Chilean Binational Center of Artificial Intelligence, ANID Strengthening R\&D capabilities Program CTI230007 Inria Chile.
This project also received funding from the BPI Scribe projects. This work was granted access to the HPC resources of IDRIS under the allocation 2025-A0180616119 made by GENCI.


\bibliography{custombib}

\appendix

\section{Dataset Details}
\label{app:data}
\subsection{Population partition and graph structure}
We use $N=100$K individuals, following 
\citet{allen-zhu_physics_2023-1}, who show that 
training on larger populations does not change 
their conclusion on knowledge storage, extraction, or 
manipulation. The $N=100$K individuals are split 
into two disjoint populations of $50$K each. 
$\mathcal{P}_{\text{comp}}$ contains individuals 
whose \texttt{friend} and \texttt{enemy} relations 
point exclusively to other members of 
$\mathcal{P}_{\text{comp}}$, forming a closed 
relational component over which compositional 
chains can be constructed. $\mathcal{P}_{\text{held}}$ 
contains individuals whose relations point 
exclusively within $\mathcal{P}_{\text{held}}$. 
This strict containment guarantees that no member 
of $\mathcal{P}_{\text{held}}$ ever appears as a 
head, bridge, or target entity in any compositional 
sequence seen during pretraining. $\mathcal{P}_{\text{comp}}$ 
is further split for finetuning: $75\%$ form 
$\mathcal{P}^{\text{train}}_{\text{comp}}$ (2-hop 
questions seen during QA finetuning) and the 
remaining $25\%$ form $\mathcal{P}^{\text{test}}_{\text{comp}}$ 
(2-hop questions held out).

\subsection{Attributes and relations}
Each individual is described by six atomic attributes following \citet{allen-zhu_physics_2023-1}: \texttt{birthday}, \texttt{birthcity}, \texttt{university}, \texttt{major}, \texttt{company}, and \texttt{workcity}. We add two directional inter-individual relations, \texttt{friend} and \texttt{enemy}, each pointing from an individual to exactly one other individual in the same population. Relations are directional: $X$ being the friend of $Y$ does not imply $Y$ being the friend of $X$.

\subsection{Augmentation format examples}
\label{app:format-examples}
Table~\ref{tab:format_examples} illustrates each augmentation format on a single running example. Atomic 1-hop biographies (shared across all conditions) use the \texttt{multi5p-permute} format of \citet{allen-zhu_physics_2023-1}. RDF sequences encode the same facts as structured triples with dedicated special tokens \texttt{[ENTITY]}, \texttt{[RELATION]}, and \texttt{[VALUE]}. Explicit formats name the bridge entity, implicit formats omit it.

\begin{table}[!h]
\centering
\scriptsize
\setlength{\tabcolsep}{3pt}
\renewcommand{\arraystretch}{1.3}
\resizebox{\linewidth}{!}{
\begin{tabular}{p{2.3cm} p{6.2cm}}
\toprule
Format & Example sequence \\
\midrule
1-hop NL (base) & \textit{Marcus Halloway was born on June 14, 1967. He studied Linguistics at Northgate University.} \\
1-hop RDF & \texttt{[ENTITY]} Marcus Halloway \texttt{[RELATION]} birthday \texttt{[VALUE]} June 14, 1967 \\
NL 2-hop implicit & \textit{Marcus Halloway's friend was born in Ashford.} \\
NL 2-hop explicit & \textit{Marcus Halloway's friend Delia Crane was born in Ashford.} \\
RDF 2-hop implicit & \texttt{[ENTITY]} Marcus Halloway \texttt{[RELATION]} friend.birthcity \texttt{[VALUE]} Ashford \\
RDF 2-hop explicit & \texttt{[ENTITY]} Marcus Halloway \texttt{[RELATION]} friend \texttt{[VALUE]} Delia Crane \texttt{[RELATION]} birthcity \texttt{[VALUE]} Ashford \\
\bottomrule
\end{tabular}
}
\caption{\textbf{Augmentation format examples.} Each format expresses the same underlying 2-hop fact (the birth city of Marcus Halloway's friend). Implicit formats omit the bridge entity (Delia Crane) and explicit formats include it.}
\label{tab:format_examples}
\end{table}

\subsection{Question counts}
Table~\ref{tab:qa_counts} reports the number of QA pairs used for finetuning and evaluation, per population and hop count. Questions for $\mathcal{P}^{\text{train}}_{\text{comp}}$ are seen during finetuning, questions for $\mathcal{P}^{\text{test}}_{\text{comp}}$ and $\mathcal{P}_{\text{held}}$ are used only at evaluation.

\begin{table}[!h]
\centering
\small
\setlength{\tabcolsep}{5pt}
\begin{tabular}{lccc}
\toprule
& $\mathcal{P}^{\text{train}}_{\text{comp}}$ & $\mathcal{P}^{\text{test}}_{\text{comp}}$ & $\mathcal{P}_{\text{held}}$ \\
\midrule
1-hop & {300K} & {100K} & {400K} \\
2-hop & {600K} & {200K} & {800K} \\
3-hop & {1.2M}& {400K} & {1.6M} \\
\bottomrule
\end{tabular}
\caption{\textbf{QA pair counts} per population and hop count. Counts scale with hop depth as the number of relational paths grows.}
\label{tab:qa_counts}
\end{table}

\section{Training and Optimization Details}
\label{app:train}

\subsection{Architecture and pretraining configuration}
We pretrain a GPT-2 small architecture (12 layers, 12 heads, 768 hidden dimension, 124M parameters) from scratch using rotary positional embeddings \citep{su_roformer_2023}. Optimization uses AdamW with peak learning rate $10^{-3}$, minimum learning rate $10^{-4}$, 1000-step warmup, cosine decay, and gradient clipping at 1.0. Training uses 512-token context windows and a batch size of 49\,152 tokens, for 800\,000 steps. We train our models on 4-H100 GPU, pretraining take around 8 hours and finetuning 2 hours. The vocabulary size $|\mathcal{V}|$ for experiments without RDF is $50,264$, and for experiments with RDF is $50304$ (a number divisible by a power of 2 to speed up training).

\subsection{Finetuning configuration}
We use LoRA \citep{hu_lora_2021} applied to query, value, and embedding layers with AdamW ($lr=3\times10^{-4}$, weight decay $0.01$). We sweep LoRA ranks $r_{\text{qv}}\in\{8,16,32\}$ and $r_{\text{emb}}\in\{32,64,128\}$ and report in \Cref{tab:aug_results} the best result per condition.

\subsection{Training statistics}
In \cref{tab:token_count}, we detail the number of unique tokens for each experiment and compare our training budget against Chinchilla scaling laws \cite{10.5555/3600270.3602446}. Following the methodology described in \cite{allen-zhu_physics_2023-1}, we perform multiple passes over the dataset until the model converges, ensuring complete memorization of the training distribution.

\begin{table}[!h]
\centering
\scriptsize
\setlength{\tabcolsep}{2pt}
\renewcommand{\arraystretch}{0.95}
\resizebox{\linewidth}{!}{
\begin{tabular}{cccc}
\toprule
\# Exp. & Unique Tokens & Total Tokens & Chinchilla Ratio \\
\midrule
Bas. GPT-S & 60 M. & 39.3 B & $\times$ 15.9 \\
Bas. GPT-M & 60 M. & 39.3 B & $\times$ 5.23  \\
Bas. GPT-L & 60 M. & 39.3 B & $\times$ 3.54  \\
\midrule
Exp. 1 & 111.4 M. & 39.3 B & $\times$ 15.9  \\
Exp. 2 & 129.6 M. & 39.3 B & $\times$ 15.9  \\
Exp. 3 & 142.1 M. & 39.3 B & $\times$ 15.9  \\
Exp. 4 & 127.8 M. & 39.3 B & $\times$ 15.9  \\
Exp. 5 & 121.4 M. & 39.3 B & $\times$ 15.9  \\
Exp. 6 & 249.2 M. & 39.3 B & $\times$ 15.9  \\
Exp. 7 & 271.0 M. & 39.3 B & $\times$ 15.9  \\
Exp. 8 & 385.1 M. & 39.3 B & $\times$ 15.9  \\
Exp. 9 & 396.4 M. & 39.3 B & $\times$ 15.9  \\
\bottomrule
\end{tabular}
}
\caption{\textbf{Token statistics.} Comparison of unique token counts, total tokens seen during training, and the compute ratio relative to Chinchilla-optimal scaling laws.}
\label{tab:token_count}
\end{table}

\subsection{Full Fine-tuning and Catastrophic Forgetting}
In the experiments presented in \Cref{tab:baseline}, we observe that the model struggles to converge, even when using the following LoRA parameters: $r_{\text{qv}}\in\{8,16,32\}$ and $r_{\text{emb}}\in\{32,64,128\}$. Furthermore, we demonstrate that full fine-tuning, by forcing the model to converge on 2-hop relations, leads to catastrophic forgetting on the $\mathcal{P}_{\text{held}}$ population.

\begin{table}[!h]
\centering
\small
\begin{tabular}{llccc}
\toprule
Fine-tuning & Pop. & 1-hop & 2-hop & 3-hop \\
\midrule
\midrule
1+2-hop    & $\mathcal{P}_{\text{comp}}$ & \valcell{1.00} & \valcell{0.99} & \valcell{0.01} \\
           & $\mathcal{P}_{\text{held}}$ & \valcell{0.01} & \valcell{0.01} & \valcell{0.01} \\
\bottomrule
\end{tabular}
\caption{\textbf{Full fine-tuning.} Full fine-tuning on 1 and 2-hop questions leads to catastrophic forgetting on 1-hop generalization.}
\label{tab:forget}
\end{table}

\begin{table}[!ht]
\centering
\small
\begin{tabular}{lccc}
\toprule
\# Exp & $\mathcal{P}^{\text{train}}_{\text{comp}}$ & $\mathcal{P}^{\text{test}}_{\text{comp}}$ & $\mathcal{P}^{\text{held}}_{\text{comp}}$ \\
\midrule
Exp 1 & \valcell{0.10} & \valcell{0.00} & \valcell{0.00} \\
Exp 2 & \valcell{0.77} & \valcell{0.49} & \valcell{0.00} \\
Exp 3 & \valcell{0.00} & \valcell{0.00} & \valcell{0.00} \\
Exp 4 & \valcell{0.92} & \valcell{0.51} & \valcell{0.00} \\
Exp 5 & \valcell{0.00} & \valcell{0.00} & \valcell{0.00} \\
Exp 6 & \valcell{0.81} & \valcell{0.50} & \valcell{0.00} \\
Exp 7 & \valcell{0.79} & \valcell{0.44} & \valcell{0.00} \\
Exp 8 & \valcell{0.81} & \valcell{0.60} & \valcell{0.00} \\
Exp 9 & \valcell{0.93} & \valcell{0.51} & \valcell{0.00} \\
\bottomrule
\end{tabular}
\caption{\textbf{Conditional analysis.} Success rate of 2-hop composition given that both constituent 1-hop sub-questions are answered correctly.}
\label{tab:condition}
\end{table}

\subsection{Data augmentation ratios}
\label{app:ratio}
Each batch mixes atomic biographies from all 
individuals with compositional augmentation 
sequences drawn exclusively from 
$\mathcal{P}_{\text{comp}}$. We fix the 
atomic/compositional ratio at 30/70, placing our 
experiments in the high-supervision regime shown 
to be necessary for implicit composition to emerge 
\citep{ ye2026transformers, 
wang_grokking_2024}. This is a deliberately 
favorable setting for composition: the absence of 
transfer to $\mathcal{P}_{\text{held}}$ under this regime cannot be attributed to insufficient 
augmentation. Multi-format conditions divide the 
compositional portion equally across constituent 
formats. Full mixing ratios are in 
\Cref{tab:train_ratios}.

\begin{table}[!h]
\centering
\scriptsize
\setlength{\tabcolsep}{2pt}
\renewcommand{\arraystretch}{0.95}
\resizebox{\linewidth}{!}{
\begin{tabular}{l p{3.2cm} l}
\toprule
\# Exp. & Setting & Ratio \\
\midrule
Exp. 1 & 1-hop RDF & 50--50 \\
Exp. 2 & NL 2-hop (imp.) & 30--70 \\
Exp. 3 & NL 2-hop (exp.) & 30--70 \\
Exp. 4 & RDF 2-hop (imp.) & 30--70 \\
Exp. 5 & RDF 2-hop (exp.) & 30--70 \\
Exp. 6 & RDF 2-hop (exp+imp) & 30--35--35 \\
Exp. 7 & NL 2-hop (exp+imp) & 30--35--35 \\
Exp. 8 & NL+RDF 2-hop (exp+imp) & 30--17.5$\times$4 \\
Exp. 9 & Full (NL+RDF+1-hop RDF) & 15--15--17.5$\times$4 \\
\bottomrule
\end{tabular}
}
\caption{\textbf{Pretraining mixtures.} Ratios indicate the proportion of atomic 1-hop biographical text and compositional augmentation sequences in each training batch. The baseline (Exp.~0) uses only \texttt{multi5p-permute} biographical text and therefore has no mixing ratio.}
\label{tab:train_ratios}
\end{table}

\section{Additional Experimental Results}
\label{app:exp}

\subsection{Full main results}
Table~\ref{tab:aug_results_full} reports 1-hop, 2-hop, and 3-hop accuracy across all 9 augmentation conditions and the baseline, for the three evaluation populations including $\mathcal{P}^{\text{train}}_{\text{comp}}$, which is omitted from the main text. Accuracy on $\mathcal{P}^{\text{train}}_{\text{comp}}$ reflects performance on 2-hop questions seen during finetuning. The gap between $\mathcal{P}^{\text{train}}_{\text{comp}}$ and $\mathcal{P}^{\text{test}}_{\text{comp}}$ isolates QA memorization from genuine compositional transfer. 2-hop accuracy on $\mathcal{P}_{\text{held}}$ remains at chance across every condition.
\begin{table}[!ht]
\centering
\small
\setlength{\tabcolsep}{6pt}
\begin{tabular}{lcc}
\toprule
Attribute & Base & Exp. 8 \\
\midrule
\texttt{birthday}  & \valcell{0.92} & \valcell{0.88} \\
\texttt{birthcity} & \valcell{0.97} & \valcell{0.94} \\
\texttt{major}     & \valcell{0.96} & \valcell{0.94} \\
\texttt{employer}  & \valcell{0.96} & \valcell{0.95} \\
\texttt{friend}    & \valcell{0.86} & \valcell{0.60} \\
\texttt{enemy}     & \valcell{0.87} & \valcell{0.60} \\
\bottomrule
\end{tabular}
\caption{\textbf{1-hop accuracy by attribute on $\mathcal{P}^{\text{test}}_{\text{comp}}$.} Scalar attributes are retrieved near-perfectly under both conditions. Relational targets (\texttt{friend}, \texttt{enemy}) are retrieved far less reliably, and degrade markedly under Exp.~8.}
\label{tab:attr_1hop}
\end{table}

\begin{table*}[!ht]
\centering
\small
\setlength{\tabcolsep}{4pt}
\resizebox{0.65\linewidth}{!}{
\begin{tabular}{cccc|ccc|ccc}
\toprule
\multirow{2}{*}{\#} &
\multicolumn{3}{c}{$\mathcal{P}^{\text{train}}_{\text{comp}}$} &
\multicolumn{3}{c}{$\mathcal{P}^{\text{test}}_{\text{comp}}$} &
\multicolumn{3}{c}{$\mathcal{P}_{\text{held}}$} \\
\cmidrule(lr){2-4} \cmidrule(lr){5-7} \cmidrule(lr){8-10}
& 1-hop & 2-hop & 3-hop & 1-hop & 2-hop & 3-hop &
1-hop & 2-hop & 3-hop \\
\midrule
Exp. 0 & \valcell{1.00} & \valcell{0.01} & \valcell{0.01} & \valcell{0.97} & \valcell{0.01} & \valcell{0.01} & \valcell{0.97} & \valcell{0.01} & \valcell{0.01} \\
\midrule
Exp. 1 & \valcell{1.00} & \valcell{0.01} & \valcell{0.01} & \valcell{0.97} & \valcell{0.01} & \valcell{0.01} & \valcell{0.97} & \valcell{0.01} & \valcell{0.01} \\
Exp. 2 & \valcell{1.00} & \valcell{0.72} & \valcell{0.01} & \valcell{0.83} & \valcell{0.54} & \valcell{0.01} & \valcell{0.75} & \valcell{0.01} & \valcell{0.01} \\
Exp. 3 & \valcell{1.00} & \valcell{0.02} & \valcell{0.02} & \valcell{0.96} & \valcell{0.02} & \valcell{0.02} & \valcell{0.89} & \valcell{0.02} & \valcell{0.02} \\
Exp. 4 & \valcell{1.00} & \valcell{0.97} & \valcell{0.01} & \valcell{0.96} & \valcell{0.71} & \valcell{0.01} & \valcell{0.40} & \valcell{0.01} & \valcell{0.01} \\
Exp. 5 & \valcell{1.00} & \valcell{0.02} & \valcell{0.02} & \valcell{0.97} & \valcell{0.02} & \valcell{0.02} & \valcell{0.37} & \valcell{0.02} & \valcell{0.02} \\
Exp. 6 & \valcell{1.00} & \valcell{0.90} & \valcell{0.01} & \valcell{0.98} & \valcell{0.69} & \valcell{0.01} & \valcell{0.48} & \valcell{0.01} & \valcell{0.01} \\
Exp. 7 & \valcell{1.00} & \valcell{0.76} & \valcell{0.01} & \valcell{0.89} & \valcell{0.68} & \valcell{0.01} & \valcell{0.79} & \valcell{0.01} & \valcell{0.01} \\
Exp. 8 & \valcell{1.00} & \valcell{0.90} & \valcell{0.01} & \valcell{0.99} & \valcell{0.76} & \valcell{0.01} & \valcell{0.82} & \valcell{0.01} & \valcell{0.01} \\
Exp. 9 & \valcell{1.00} & \valcell{0.98} & \valcell{0.01} & \valcell{0.98} & \valcell{0.71} & \valcell{0.01} & \valcell{0.80} & \valcell{0.01} & \valcell{0.01} \\
\bottomrule
\end{tabular}
}
\caption{\textbf{Full results across augmentation conditions.} Exact-match accuracy for 1-hop, 2-hop, and 3-hop queries on $\mathcal{P}^{\text{train}}_{\text{comp}}$ (2-hop questions seen at finetuning), $\mathcal{P}^{\text{test}}_{\text{comp}}$ (2-hop questions held out from finetuning), and $\mathcal{P}_{\text{held}}$ (never compositionally exposed during pretraining). Each value is the mean over 5 finetuning and evaluation seeds; standard deviations are below $0.006$ for all cells. Row 0 is the baseline. Cell color encodes accuracy from low (purple) to high (green). The high accuracy on $\mathcal{P}^{\text{train}}_{\text{comp}}$ together with the accuracy retained on $\mathcal{P}^{\text{test}}_{\text{comp}}$ for the strongest conditions indicates transfer across unseen questions. The persistent chance-level accuracy on $\mathcal{P}_{\text{held}}$ indicates no transfer across populations.}
\label{tab:aug_results_full}
\end{table*}
\subsection{Validation on an Llama-based architecture}
\label{app:llama}
To validate that the compositionality gap is not an artifact of the GPT-2 architecture, we replicate the full pipeline on a parameter-matched Llama-style model~\citep{grattafiori2024llama3herdmodels}. We keep the GPT-2 Small topology (12 layers, 12 heads,$d_{\text{model}}=768$, context $512$) and swap every GPT-2 specific component for its Llama counterpart: RMSNorm ($\varepsilon=10^{-5}$) instead of LayerNorm, a SwiGLU feed-forward block (hidden dimension $2048 = \tfrac{8}{3}d_{\text{model}}$ rounded to a multiple of $256$, against $4d_{\text{model}}=3072$ for the GPT-2 MLP) instead of the GELU MLP, RoPE applied to the full head dimension rather than to $25\%$ of it, and no bias in any linear layer. Attention stays multi-head ($n_{\text{kv}}=n_{\text{heads}}$, no GQA). We keep the GPT-2 BPE tokeniser so both models consume the same token stream, and tie the input and output embeddings (untying, as in stock Llama-3, would add $38.6$M parameters at this vocabulary size); the model has $123.55$M parameters against $123.65$M for GPT-2 Small. Pretraining uses the same mixtures, optimiser and schedule for $500$K steps, and finetuning uses LoRA with a fixed $(r_{qv},r_{\text{emb}})=(8,32)$ for $150$K steps. \cref{tab:aug_results_full_llama} reproduces our findings: several augmentations induce composition on $\mathcal{P}_{\text{comp}}$ (Exp.~2, 4, 6, 7), yet no condition yields any generalisation on $\mathcal{P}_{\text{held}}$. Numbers are exact-match accuracy on all populations under greedy decoding, and average across 5 different finetuning seed and a fixed LoRA rank.

\subsection{LoRA rank sweep}
Table~\ref{tab:lora_appendix} reports the full LoRA rank sweep on $\mathcal{P}^{\text{train}}_{\text{comp}}$. Each cell gives 1-hop / 2-hop first-token accuracy for a given $(r_{\text{qv}}, r_{\text{emb}})$ pair. Increasing LoRA rank yields only marginal variation in 2-hop accuracy within each condition, confirming that the compositional outcome is determined by the pretraining augmentation rather than by finetuning capacity. Main-text results use the best $(r_{\text{qv}}, r_{\text{emb}})$ pair per condition.

\begin{table*}[!h]
\centering
\small
\setlength{\tabcolsep}{3pt}
\resizebox{0.95\linewidth}{!}{
\begin{tabular}{lccc|ccc|ccc}
\toprule
& \multicolumn{3}{c|}{Exp. 1}
& \multicolumn{3}{c|}{Exp. 2}
& \multicolumn{3}{c}{Exp. 3} \\
\cmidrule(lr){2-4} \cmidrule(lr){5-7} \cmidrule(lr){8-10}
$r_{\text{qv}}$ / $r_{\text{emb}}$
& 32 & 64 & 128
& 32 & 64 & 128
& 32 & 64 & 128 \\
\midrule
8  & 1.0/0.01 & 1.0/0.01 & 1.0/0.01
   & 1.0/0.60 & 1.0/0.63 & 1.0/0.65
   & \textbf{1.0/0.02} & 1.0/0.01 & 1.0/0.01 \\
16 & 1.0/0.01 & 1.0/0.01 & 1.0/0.01
   & 1.0/0.66 & 1.0/0.69 & 1.0/0.70
   & 1.0/0.01 & 0.99/0.01 & 0.99/0.02 \\
32 & 1.0/0.01 & 1.0/0.01 & 1.0/0.01
   & \textbf{1.0/0.72} & 1.0/0.71 & 1.0/0.69
   & 0.99/0.01 & 0.98/0.01 & 0.98/0.01 \\
\midrule
& \multicolumn{3}{c|}{Exp. 4}
& \multicolumn{3}{c|}{Exp. 5}
& \multicolumn{3}{c}{Exp. 6} \\
\midrule
8  & 1.0/0.88 & 1.0/0.91 & 1.0/0.92
   & 1.0/0.02 & \textbf{1.0/0.03} & 1.0/0.02
   & 1.0/0.81 & 1.0/0.83 & 1.0/0.85 \\
16 & 1.0/0.93 & 1.0/0.95 & 1.0/0.96
   & 1.0/0.01 & 1.0/0.02 & 1.0/0.01
   & \textbf{1.0/0.90} & 1.0/0.89 & 1.0/0.87 \\
32 & \textbf{1.0/0.97} & 1.0/0.96 & 1.0/0.94
   & 1.0/0.01 & 1.0/0.01 & 0.99/0.01
   & 1.0/0.88 & 1.0/0.86 & 1.0/0.85 \\
\midrule
& \multicolumn{3}{c|}{Exp. 7}
& \multicolumn{3}{c|}{Exp. 8}
& \multicolumn{3}{c}{Exp. 9} \\
\midrule
8  & 1.0/0.71 & 1.0/0.73 & 1.0/0.74
   & 1.0/0.87 & \textbf{1.0/0.90} & 1.0/0.88
   & 1.0/0.90 & 1.0/0.92 & 1.0/0.94 \\
16 & 1.0/0.73 & 1.0/0.75 & 1.0/0.74
   & 1.0/0.85 & 1.0/0.86 & 1.0/0.84
   & 1.0/0.95 & 1.0/0.96 & 1.0/0.97 \\
32 & \textbf{1.0/0.76} & 1.0/0.75 & 1.0/0.72
   & 1.0/0.82 & 1.0/0.83 & 1.0/0.81
   & 1.0/0.97 & \textbf{1.0/0.98} & 1.0/0.96 \\
\bottomrule
\end{tabular}
}
\caption{\textbf{LoRA rank sweep on $\mathcal{P}^{\text{train}}_{\text{comp}}$.} Each cell reports 1-hop / 2-hop exact-match accuracy for a $(r_{\text{qv}}, r_{\text{emb}})$ pair. Best 2-hop configuration per condition in bold. Variation across ranks is moderate and does not change the ranking of conditions, indicating that finetuning capacity is not the limiting factor.}
\label{tab:lora_appendix}
\end{table*}

\subsection{Per-attribute results}
\label{app:attribute_res}

Scalar attributes are retrieved near-perfectly under both conditions. Relational targets behave differently: \texttt{friend} and \texttt{enemy} drop from $\sim$0.86 at baseline to $0.60$ under Exp.~8. The two cases are not symmetric. A scalar attribute is drawn from a small, low-cardinality set, whereas a relational target is one specific individual among 100K, identified by a name. Retrieving it requires discriminating that name from a very large space of similar tokens, where first names recur across many individuals and offer little disambiguating signal. Under Exp.~8, the compositional augmentation dilutes the atomic biographical text in each batch (\Cref{app:ratio}), reducing exposure to exactly the per-individual name associations that relational retrieval depends on. High-cardinality name retrieval is therefore the first capability to degrade when atomic supervision is diluted, while low-cardinality scalar attributes remain robust.

\vspace{3 pt}
\subsection{Conditional Analysis}
In \Cref{tab:condition}, we address the following question: \textit{"Among instances where both 1-hop sub-questions are answered correctly, what fraction results in a correct 2-hop prediction?"} 

Given that our entity bridge is always an individual within a set of 100,000, the task requires precise retrieval. While the overall accuracy on $\mathcal{P}_\text{held}$ reaches 97\% for the baselines (\Cref{tab:baseline}) and 83\% for Exp. 8, we observe a catastrophic performance drop in compositional reasoning. For instance, consider the query: \textit{"What is the birthday of X's friend?"} Even when the model correctly identifies the friend ($Y$) and retrieves $Y$'s birthday, it alwats fails to compose these facts for individuals in $\mathcal{P}_{\text{held}}$. As shown in \Cref{tab:condition}, despite the guarantee of perfect 1-hop awnsers, the composition success rate remains remarkably low across most experimental settings.

\section{Ethics/Transparency Statement}

We used AI-assisted tools for language polishing and proofreading of the manuscript. The authors have reviewed, edited, and approved the final content.
\end{document}